\documentclass[letterpaper, 10 pt, conference]{ieeeconf}  

\IEEEoverridecommandlockouts                              

\usepackage{xcolor}
\usepackage[dvipdfmx]{graphicx}
\usepackage{comment}
\usepackage{nidanfloat}
\usepackage{multirow} 

\title{\LARGE \bf Achieving Faster and More Accurate Operation \\ of Deep Predictive Learning}

\author{Masaki Yoshikawa$^{1}$, Hiroshi Ito$^{1}$, Tetsuya Ogata$^{1}$
\thanks{$^{1}$ All authors is with Department of Intermedia Art and Science School of Fundamental Science and Engineering, Waseda University, Tokyo, 169-855, Japan. {\tt\small ogata@waseda.jp}}
}

\begin{document}

\maketitle
\thispagestyle{empty}
\pagestyle{empty}

\begin{figure*}[!b]
    \centering
    \includegraphics[width=\linewidth]{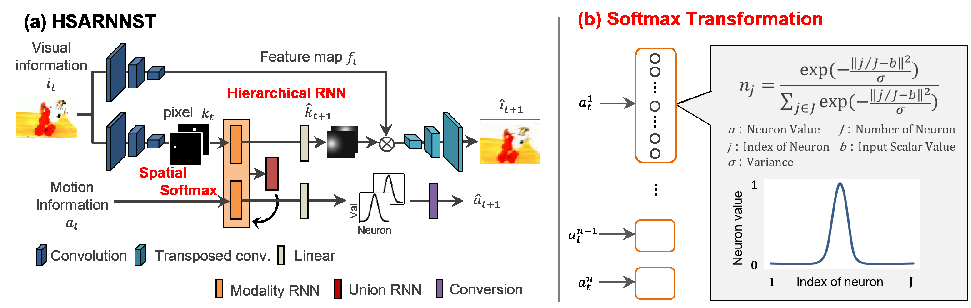}
    \caption{Overview of model and data processing. (a) Network structure of HSARNNST. (b) Data processing of Softmax Transformation.}
    \label{fig:model_overview}
\end{figure*}

\section{Introduction}
One of the key challenges for social implementation of robots is to achieve both high speed and high precision in operation.
Robots used in factories are excellent at quickly and accurately reproducing predefined actions such as assembling and transporting products, but they cannot handle tasks that require environmentally specific actions, such as cleaning and cooking.
To solve this problem, research on robot applications of deep learning has been attracting attention.
By performing end-to-end learning with sensor data as the robot interacts with its environment, it can autonomously execute desired behaviors without the need to program any trajectory plans. By collecting and learning from a large dataset, RT-1\cite{brohan2022rt} is capable of performing more than 700 tasks in everyday life environments that would be difficult to achieve in the past.
A type of mimetic learning, ACT\cite{zhao2023learning} uses teleoperation to teach motions and thus can perform multiple tasks, including twin-arm motions and movements, at a reduced cost.

However, many machine learning-based motion generation methods have issues with model inference speed and hand position estimation accuracy.
To achieve both high speed and high accuracy in motion, two issues need to be cleared.
The first is high-quality training data. The inference accuracy of machine learning models is greatly influenced by the quality of the instruction data.
In particular, in order for a robot to achieve high-speed motion in a mimetic learning framework, high-speed motion teaching by the operator is indispensable, and issues remain regarding hardware devices and teaching methods.
The second is the inference speed of the model. While large-scale models can perform a wide variety of actions, they lack real-time performance because the models must be inferred over a network.
Even if fast inference can be performed, it is extremely sensitive to changes in sensor data and the external environment, so a mechanism that enables stable and smooth motion inference is needed.

In this paper, we propose a motion generation model focused on high-speed, high-precision tasks, using a Sport Stacking task as an example.
In order to collect high-quality data, motion is taught slowly over 40 seconds per task.
When generating actual motions, we achieve high-speed motion generation by having the model infer at several times the speed of the real world.
The results of having the model infer at three times the speed during teaching using a real robot show that it can stack randomly placed cups with an average success rate of 94\%.

\section{Method}
Fig.\ref{fig:model_overview} (a) shows an overview of the motion generation model, which consists of three features.
The first is a visual attention mechanism that is robust to changes in object position.
An image $i_t$ at time $t$ is input to the Convolutioanl Neural Network, and by applying Spatial Softmax\cite{finn2016deep} to the extracted image features, important regions in the image are extracted as pixel values $k_t$.
The second is a hierarchical RNN that reduces visual noise during fast inference.
It consists of Modality RNN that learns time series of each modality independently and Union RNN that performs integrated learning of them.
One Modality RNN receives visual information $k_t$ and the other receives motion information $a_t$, and then each Modality RNN predicts their values at the next time 
 ($\hat{k}_{t+1}, \hat{a}_{t+1}$) independently.
The hidden state of each Modality RNN is input to Union RNN, and the predicted internal state at the next time is fed back to Modality RNN.
After extracting features for each Modality, Union RNN performs integrated learning, which enables balanced learning of visual and motion information. In addition, learning once abstracted information enables noise-robust inference compared to the usual LSTM that directly updates the internal state from the input at each time.
The third is the decoder trained with Softmax Transformation, which enables accurate and stable state prediction.
Normally, a single neuron is used to make a single posture prediction, but this has the problems of low expressiveness and excessive response to input values.
In this paper, each element of predicted motion information $\hat{a}_{t+1}$ is output via a discrete probability distribution using 2000 neurons.
This enables stable and accurate prediction.
Training is performed on the basis of the teacher data generated using Softmax Transformation.
Fig.\ref{fig:model_overview}(b) shows the Softmax Transformation of motion information. During inference, the predicted motion information $\hat{a}_{t+1}$ converted from the largest index of 2000 neurons is input to the robot.

\section{Experiment and Results}
\begin{figure*}[!t]
    \centering
    \includegraphics[width=\linewidth]{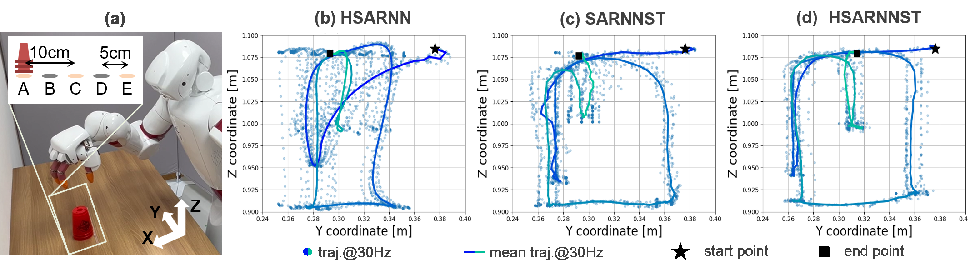}
    \caption{Experimental environment and results. (a) Robot performs the task of stacking cups in five positions, including an untaught position. (b)(c)(d) Hand trajectories of the compared models during motion generation at position C.}
    \label{fig:exp_results}
\end{figure*}

Fig.\ref{fig:exp_results}(a) shows an overview of the experimental setup.
We verified the effectiveness of the proposed method using an example of a Sport Stacking task.
We taught the robot to grasp three overlapping cups and stack them on two levels using a Joystick controller.
Three positions in a straight line at 10 cm intervals were set as teaching positions, and the positions between them were set as untaught positions.
Data was collected once for each teaching position for 40 seconds at 10 Hz (400 steps).

To investigate the effect of the presence or absence of Hierarchical RNN and Softmax Transformation on the task success rate, we performed comparative experiments with four different models (SARNN, HSARNN, SARNNST, and HSARNNST). SARNN is the base model, which consists of only one LSTM and directly predicts the motion information $\hat{a}_{t+1}$ without probability distribution. When a Hierarchical RNN is used, the model name starts with "H", and when Softmax Transformation is used, it ends with "ST".
Table \ref{tab:success_rate} shows the task success rate when the motion is generated at three times the speed of teaching (30 Hz) and the proposed method has the highest average success rate.
To investigate the reason for the difference in task success rates, Fig.\ref{fig:exp_results}(b)(c)(d) show the hand trajectories in the YZ plane during task execution.
The dotted line in each figure shows the hand trajectories for the 10 trials of the task with the cup positioned at C, and the solid line shows the average trajectory.
This result shows the following two things.
(1) The variation in hand position in Fig.  \ref{fig:exp_results}(b) is large, while the variation in (c) and (d) is small, indicating that the Softmax Transformation can be used to reduce the variation during inference.
(2) The trajectory in (d) is more regular compared to Fig. \ref{fig:exp_results}(c), which means that task-specific motion can be learned by the Hierarchical RNN.

\begin{table}[th]
\centering
\caption{Success Rates of the Sport Stacking Task at 30Hz [\%]}
\begin{tabular}{l|ccccc|c}
         & {\color[HTML]{3531FF} A}   & B   & {\color[HTML]{3531FF} C}  & D   & {\color[HTML]{3531FF} E}   & Ave.\\ \hline \hline
(1) SARNN    & {\color[HTML]{3531FF} 0}   & 0   & {\color[HTML]{3531FF} 0}  & 0   & {\color[HTML]{3531FF} 0}   & 0\\
(2) HSARNN   & {\color[HTML]{3531FF} 100} & 30  & {\color[HTML]{3531FF} 0}  & 30  & {\color[HTML]{3531FF} 100} & 52\\
(3) SARNNST  & {\color[HTML]{3531FF} 100} & 90  & {\color[HTML]{3531FF} 40} & 90  & {\color[HTML]{3531FF} 70}  & 78 \\
\bf{(4) HSARNNST} & \bf{{\color[HTML]{3531FF} 100}} & \bf{100} & \bf{{\color[HTML]{3531FF} 90}} & \bf{100} & \bf{{\color[HTML]{3531FF} 80}}  & \bf{94}\\
\end{tabular}
\label{tab:success_rate}
\end{table}

\section*{Acknowledgement}
This work was supported by JST [Moonshot R\&D] GrantNumber [JPMJMS2031].

\bibliographystyle{IEEEtran}
\bibliography{ref}
\end{document}